\documentclass[letterpaper, 10 pt, conference]{ieeeconf}  % Comment this line out if you need a4paper

\IEEEoverridecommandlockouts                              % This command is only needed if 
\usepackage{graphicx}
\usepackage{bm}
\usepackage[linesnumbered,lined,algoruled]{algorithm2e}
\usepackage{booktabs}
\usepackage{multirow}
\usepackage{cite}
\usepackage{fontawesome}
\usepackage{amsmath} % assumes amsmath package installed
\usepackage{amssymb}  % assumes amsmath package installed
\usepackage{float}
\UseRawInputEncoding

\title{\LARGE \bf
Collision Avoidance for Multiple UAVs in Unknown Scenarios with Causal Representation Disentanglement
}

\author{Jiafan Zhuang$^{1}$, Zihao Xia$^{1}$, Gaofei Han$^{1}$, Boxi Wang$^{1}$, Wenji Li$^{1}$, Dongliang Wang$^{1}$ \\ Zhifeng Hao$^{1}$, Ruichu Cai$^{2}$ and Zhun Fan\textsuperscript{3,4, \faEnvelopeO}
\thanks{\textsuperscript{\faEnvelopeO} Corresponding author}
\thanks{*This work is supported in part by the National Science and Technology Major Project (grant numbers 2021ZD0111501, 2021ZD0111502), the National Natural Science Foundation of China (grant numbers 62176147, 51907112, U2066212, 61961036, 62162054), Science and Technology Planning Project of Guangdong Province of China (grant numbers 2023A1515011574, 2022A1515110566, 2022A1515110660), and the STU Scientific Research Foundation for Talents (grant numbers NTF21001, NTF21052, NTF22030).}
\thanks{$^{1}$College of Engineering, Shantou University}
\thanks{$^{2}$School of Computer Science, Guangdong University of Technology}
\thanks{$^{3}$University of Electronic Science and Technology of China}
\thanks{$^{4}$International Cooperation Base of Evolutionary Intelligence and Robotics}
}

\begin{document}

\newcommand{\etal}{\textit{et al}.}
\newcommand{\ie}{\textit{i}.\textit{e}.}
\newcommand{\eg}{\textit{e}.\textit{g}.}
\newcommand{\etc}{\textit{etc}}

\maketitle
\thispagestyle{empty}
\pagestyle{empty}

%%%%%%%%%%%%%%%%%%%%%%%%%%%%%%%%%%%%%%%%%%%%%%%%%%%%%%%%%%%%%%%%%%%%%%%%%%%%%%%%
\begin{abstract}

Deep reinforcement learning (DRL) has achieved remarkable progress in online path planning tasks for multi-UAV systems. 
However, existing DRL-based methods often suffer from performance degradation when tackling unseen scenarios, since the non-causal factors in visual representations adversely affect policy learning.
To address this issue, we propose a novel representation learning approach, \ie, causal representation disentanglement, which can identify the causal and non-causal factors in representations.
After that, we only pass causal factors for subsequent policy learning and thus explicitly eliminate the influence of non-causal factors, which effectively improves the generalization ability of DRL models.
Experimental results show that our proposed method can achieve robust navigation performance and effective collision avoidance especially in unseen scenarios, which significantly outperforms existing SOTA algorithms.

\end{abstract}

%%%%%%%%%%%%%%%%%%%%%%%%%%%%%%%%%%%%%%%%%%%%%%%%%%%%%%%%%%%%%%%%%%%%%%%%%%%%%%%%
\section{INTRODUCTION}

The burgeoning unmanned aerial vehicles (UAVs) navigation technology has attracted broad attention in the fields of robotics and artificial intelligence. This technology provides a flexible, cost-effective and efficient solution for applications, such as precision agriculture~\cite{moradi2022uav,velusamy2021unmanned},  search and rescue~\cite{zhang2018analysis,ashour2023applications}, and wildlife conservation~\cite{bondi2018near}. 
To ensure an effective collaboration of multiple UAVs, the collision avoidance capability is fundamental and crucial. 
This requires each UAV to be able to identify the optimal path from the starting point to the target point while avoiding obstacles. 
Therefore, multi-UAV collision avoidance is an important task and greatly valued by researchers.

Traditional UAV collision avoidance navigation techniques, exemplified by centralized approaches~\cite{zheng2019uav}, operate under the assumption that a ground control station can communicate with all UAVs. This station can access global information about UAVs and workplace, and then generate control commands. However, the excessive reliance on communication undermines its ability to generalize to complex scenarios~\cite{wang2023blockchain}. 
In response to this challenge, some researchers have proposed decentralized UAV collision avoidance systems~\cite{li2023vg}. In these systems, each UAV independently and autonomously plans optimal path using its own sensors. 
Nonetheless, traditional path planning algorithms commonly possess a multitude of adjustable parameters, \eg, model predictive control approach~\cite{keviczky2004study}. 
The involved tedious parameter tuning according to deployment scenes also constrains the system's ability to generalize to complex and unknown scenarios~\cite{arul2020dcad,luis2019trajectory,rosmann2017integrated}.

\begin{figure}[t]
    \centering
    \includegraphics[width=245pt]{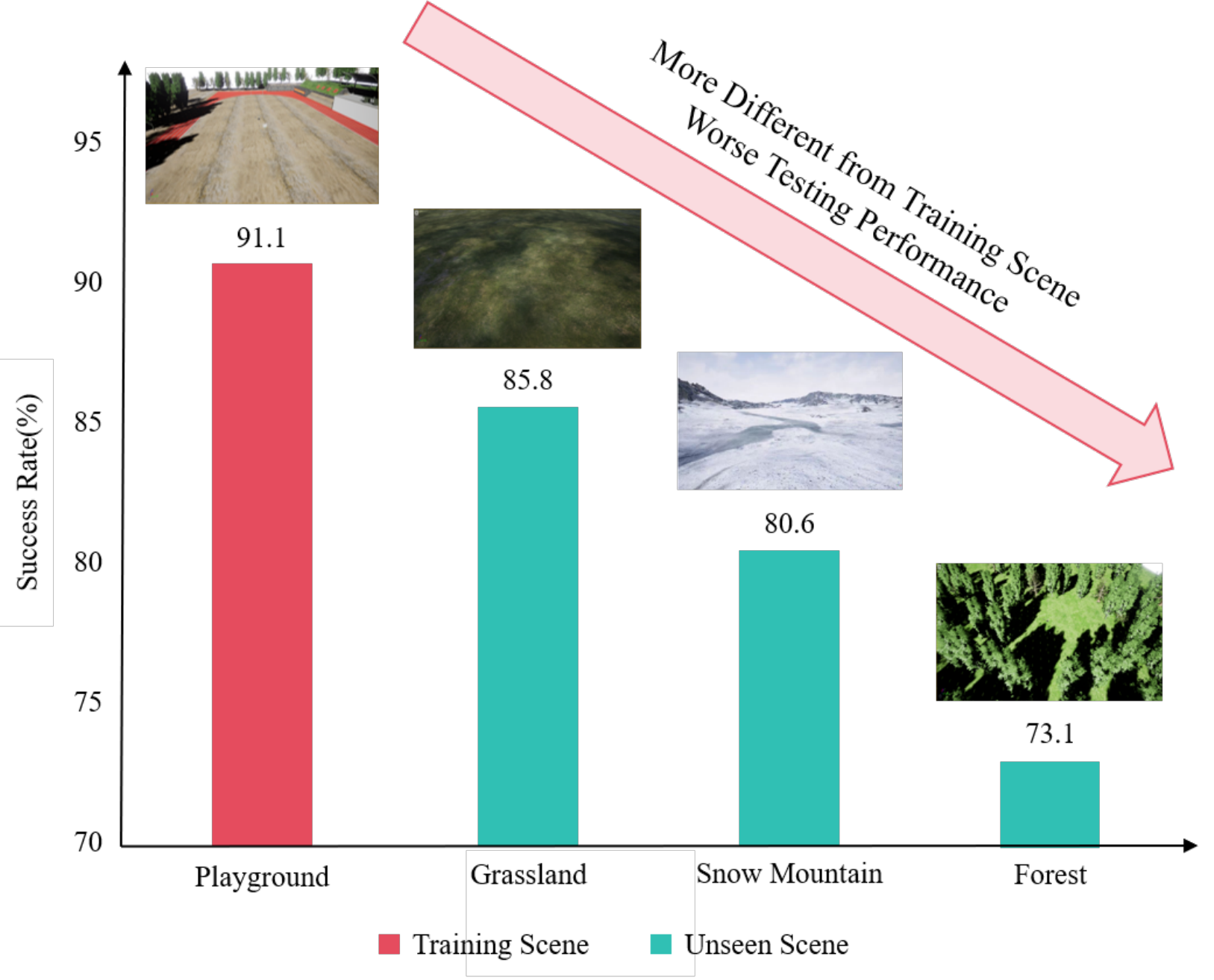}
    \caption{\textbf{The illustration of generalization ability analysis.} When facing unseen scenarios, the current DRL method will suffer severe performance degradation in navigation success rate.}
    \label{fig:analysis}
    \vspace{-4mm}
\end{figure}

With the continuous advancements in image processing and deep learning techniques in the field of robotics, a new and promising technique has attracted researchers' interest, \ie, DRL~\cite{mnih2015human,wang2019autonomous,hu2023efficient}.
Benefited from the profound feature extraction capabilities of deep neural networks, DRL can directly process high-dimensional perceptual inputs, \eg, images. Concurrently, DRL models facilitate end-to-end learning from raw inputs to policy outputs, significantly reducing the necessity for manual feature engineering design~\cite{huang2022vision}.

Similar to existing deep learning techniques, DRL is essentially a data-driven method, which assumes that the training and testing data are sampled from independent-and-identically distribution (IID) scenarios. 
However, this assumption is difficult to hold in real-world applications. For example, after training in a pre-defined scene, the DRL-based multi-UAV system may be deployed in various unseen scenes according to requirements, \eg, forests or mountain areas.

To study the generalization issue of DRL in unknown scenarios, in this work, we revisit the pioneering work of DRL-based multi-UAV collision avoidance, \ie, SAC+RAE~\cite{huang2022vision}. We provide an analysis by training the SAC+RAE model in a specific scene (\ie, playground) while testing in several different scenes to evaluate its generalization ability. As shown in Fig.~\ref{fig:analysis}, when tested in playground that is identical to training scene, the model can achieve promising performance. However, when tested in unseen scenes (\eg, grassland, snow mountain and forest), the performance significantly deteriorated. Besides, the greater the difference between the testing scenario and the training scenario, the worse the performance. Therefore, the results clearly show that the generalization issue exists in previous works.

To aid the applications of UAV techniques, there is an urgent need to develop a multi-UAV collision avoidance method that can adapt to unknown scenarios.
After studying the structure of SAC+RAE, we find that the cause of weak generalization ability of DRL may lie in the unstable and error-prone visual representations. 
Specifically, SAC+RAE uses a regularized auto-encoder (RAE)~\cite{ghosh2019variational} to learn representations of captured images in UAVs, which tends to encode all visual elements into the representations, regardless of whether it is related to the collision avoidance task or not. 
However, there exist some elements that are specific to training scenes but irrelevant to the task, \eg, background features, which would contribute to spurious correlations between perceptual inputs and actions.
When these elements change in testing scenes, it will be harmful for the model to output proper actions, thus limiting its generalization ability~\cite{zhang2020invariant,scholkopf2021toward,song2019observational}.

\begin{figure}
    \centering
    \includegraphics[width=200pt]{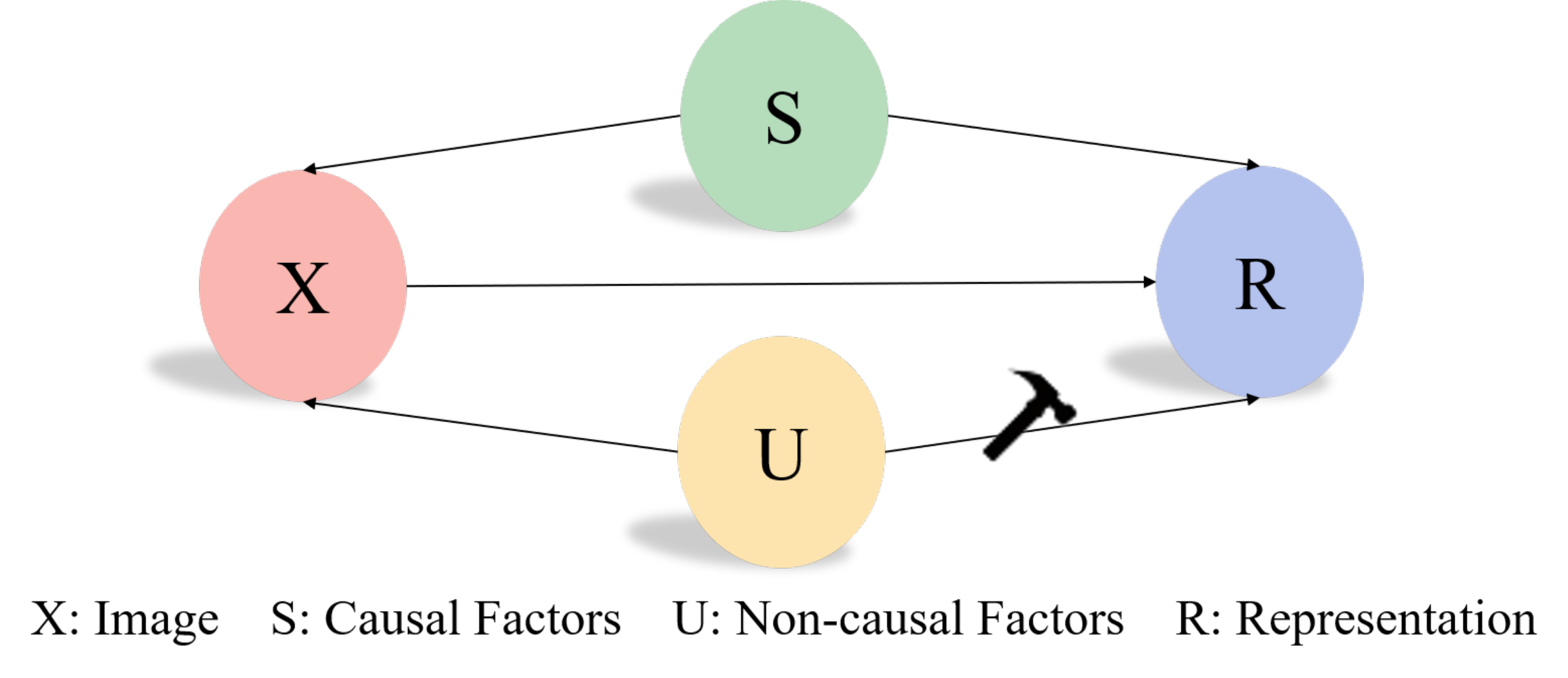}
    \caption{\textbf{Structural causal model (SCM) for representation learning.} Image X consists of causal factors S and non-causal factors U, only causal factors S has a causal impact on the representation learning process.}
    \label{fig:scm}
    \vspace{-4mm}
\end{figure}

Since the visual representations would inevitably contain task-irrelevant factors that would be harmful to subsequent policy learning, the key to improve generalization ability of DRL is to identify the involved task-irrelevant factors and discard them. 
To this aim, causal representation learning~\cite{zhang2020learning,li2024subspace} provides a feasible solution, which enables the identification and separation of causal factors affecting UAV navigation. Here, causal factors represent those can provide useful information for navigation task.
Causal representation learning can assist DRL models in better understanding the underlying causal structures across different scenarios, thereby enhancing the model's generalization capabilities when encountering unseen data or environments.
Specifically, we first construct a structural casual model (SCM) to formally model the casual relationship in representation learning, as shown in Fig.~\ref{fig:scm}. 
Here, it is hypothesized that the representation learning process (mapping from image $X$ to representation $R$) is influenced by both causal factors $S$ (\eg, obstacles features) and non-causal factors $U$ (\eg, background features), where only the former has a causal impact on the decision-making process of collision avoidance.
To identify the non-causal factors, in this work, we propose a causal representation disentanglement technique. Specifically, we design a background intervention module to provide perturbations on backgrounds and then impose different supervision signals on sub-features for disentanglement, which facilitates the separation of causal and non-causal factors. After that, only the causal sub-feature would be passed to subsequent policy learning and the effect of non-causal factors are explicitly removed, which can effectively improve the generalization ability of DRL in unknown scenarios.

Our work contributes as follows:
\begin{itemize}

\item We first study and address the generalization issue in DRL-based multi-UAV collision avoidance system from a causality perspective and propose a causal representation disentanglement framework to learn robust and causal visual representations.
\item We design a background intervention module to effectively provide perturbations on backgrounds, which facilitates the separation of causal and non-causal factors. Besides, several loss functions are proposed to guide sub-features to learn specific semantic concepts as expected.
\item We construct three typical testing scenarios for generalization study and conduct extensive experiments and analysis, which demonstrates the superiority and effectiveness of our proposed method.
\end{itemize}

\begin{figure*}
    \begin{center}
        \includegraphics[width=480pt]{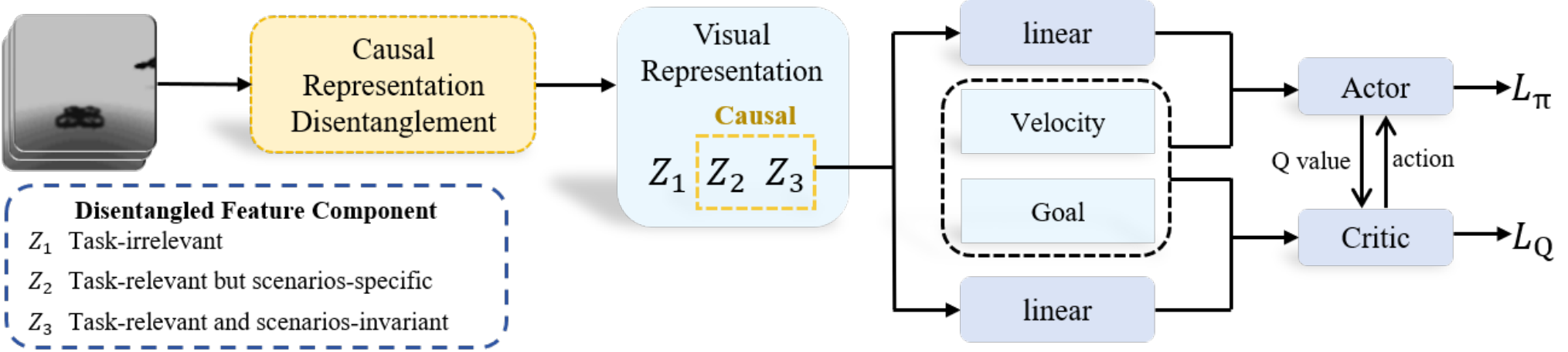}
    \end{center}
    %\vspace{2mm}
    \caption{\textbf{The architecture of our framework for Multi-UAV collision avoidance.}
    The framework follows the SAC paradigm, which takes depth images, current velocity, and relative goal position as input and outputs flight control actions. We propose a causal representation disentanglement method to optimize the visual representation extraction and only pass the causal components for subsequent policy learning.}
    \vspace{-2mm}
    \label{fig:overview}
\end{figure*}

\section{RELATED WORK}

\subsection{Multi-UAV Collision Avoidance}

Multi-UAV collision avoidance navigation is a complex field where traditional decentralized navigation methods are mostly based on the concept of velocity obstacles~\cite{van2011reciproca}. 
The classic optimal reciprocal collision avoidance (ORCA) algorithm~\cite{van2011reciprocal} derives a set of sufficient conditions for collision-free motion among multiple robots in dense and complex three-dimensional environments, enabling the rapid calculation of non-collision movements for all robots. 
Building on this foundation, Dergachev~\etal~\cite{dergachev2021distributed} integrated multi-agent path finding with ORCA, which improves the algorithm's efficiency in the tight passages or confined spaces.
However, these methods assume that perfect environmental perception by each robot is given, which is impractical in real-world applications.
Meanwhile, some researchers have explored algorithms based on imitation learning. For instance, Karnan~\etal~\cite{karnan2022voila} developed a vision-based autonomous navigation algorithm that not only mimics expert behavior but also learns strategies that can be generalized to new environments. 
However, the dependency on the quality and diversity of training data limits the generalization ability of imitation learning.
Deep reinforcement learning (DRL) techniques have proven to be effective in the field of multi-agent collision avoidance navigation, because they can learn decision-making strategies directly from sensory inputs without any prior knowledge.
In this context, Qie~\etal~\cite{qie2019joint} applied the multi-agent deep deterministic policy gradient (MADDPG) algorithm to successfully address the path planning and target allocation problems of multiple UAVs. 
Furthermore, Xue~\etal~\cite{xue2023multi} proposed a Multi-Agent Recursive Deterministic Policy Gradient (MARDPG) algorithm based on deep deterministic policy gradients to control the navigation of multiple UAVs, further demonstrating the potential of deep learning methods in solving complex navigation tasks.
However, due to the limitations in feature representation, current DRL techniques struggle to capture key features that are critical to the task at hand, thereby constraining the model's generalization capability.

\subsection{Feature Disentanglement}

The pursuit of learning interpretable feature representations constitutes a vibrant area of inquiry within the domains of computer vision and machine learning.
Aiming for the comprehension and depiction of underlying explanatory factors in a compact representation, feature disentanglement endeavors to create a joint latent feature space. In this space, chosen feature dimensions are designed to convey specific semantic information~\cite{xie2018learning}.
For example, Liu~\etal~\cite{liu2018unified} developed a unified feature disentanglement network (UFDN), which extracts invariant feature representations from multiple source domains through adversarial training. 
Peng~\etal~\cite{peng2019domain}, on the other hand, designed an innovative deep adversarial disentanglement auto-encoder (DADA) that can effectively disentangle features with just one source domain and several unlabeled target domains.
However, these approaches mainly focus on superficial correlations and do not delve into the underlying causal relationships, which limits their reliability.
To address this issue, Cai~\etal~\cite{cai2019learning} proposed to separate features into variant and invariant parts using a variational auto-encoder.
Kong~\etal~\cite{kong2022partial} approached from the perspective of data generation models, focusing on the minimal changes in cross-domain causal mechanisms, thereby proving that latent variables are partially identifiable under certain conditions. 
These conditions include a sufficient number of domains, monotonic transformations of latent variables, and invariant label distributions. 
Meanwhile, starting from causal generative mechanisms, Li~\etal~\cite{li2024subspace} demonstrated the complete identifiability of latent variables with fewer domain assumptions.

To tackle the generalization issue of DRL-based multi-UAV collision avoidance, inspired by the above studies, we design a causal representation disentanglement approach to explictly filter out the non-causal factors in visual representations.

\section{APPROACH}
\subsection{System Model and Problem Formulation}

Building on previous work SAC+RAE~\cite{huang2022vision}, we adopt a centralized training and distributed execution approach to train our policy.
In the setting of multi-UAV collision avoidance, each UAV can only avoid collisions through its sensors, \ie, a front-central camera and an inertial measurement unit, which means that the environment is only partially observable. Therefore, this setting represents a decentralized partially observable Markov decision process (Dec-POMDP). 

\begin{figure*}
    \begin{center}
        \includegraphics[width=480pt]{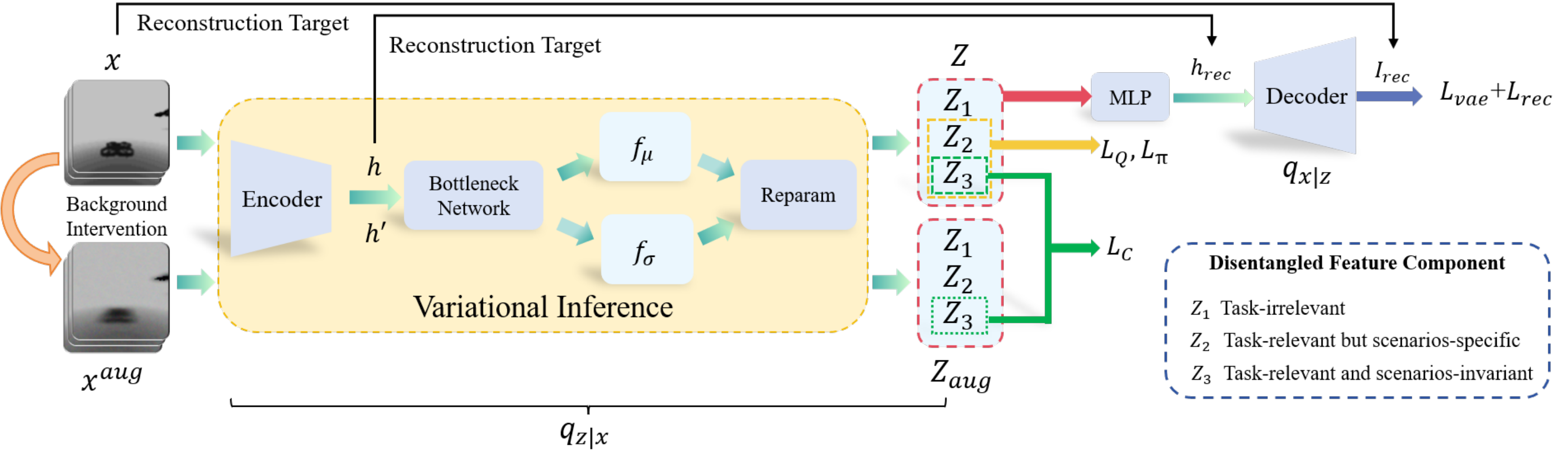}
    \end{center}
    %\vspace{4mm}
    \caption{\textbf{The illustration of causal representation disentanglement.} We use a variational auto-encoder for representation extraction, and then design specific loss function to guide pre-defined components to learn different semantic concepts. 'Reparam' represents the reparametrisation trick.}
    \vspace{-2mm}
    \label{fig:crd}
\end{figure*}

\subsubsection{Observation space}

The observation for each UAV comprises three components: the latent representation of depth images ${o}^{t}$, the relative position of the target within the body coordinate system ${o}^{g}$, and the velocity of the UAV at the current moment ${o}^{v}$.

\subsubsection{Action space}
To enhance the diversity and controllability of UAV flight, our action space is continuous. 
The action $a=[{v}_{x}^{cmd}, {v}_{z}^{cmd}, {v}_{\omega}^{cmd}]$ can be calculated from the policy network $\pi(s)$. 
Here, ${v}_{x}^{cmd}$ represents the UAV's forward velocity, indicating its speed along the front axis of the UAV's body coordinate system. 
Similarly, ${v}_{z}^{cmd}$ denotes the climbing velocity, which reflects the UAV's vertical movement.
Lastly, ${v}_{\omega}^{cmd}$ is indicative of the steering velocity, corresponding to the UAV's rotational movement around its vertical axis.

\subsubsection{Reward function}
Selecting an appropriate reward function is crucial for providing appropriate feedback to each agent. Our reward function is composed of two parts: one part, $r_g$ guides the UAV towards its target, and the other part, $r_c$, directs the UAV to avoid obstacles:
\begin{equation}
    {r}={r}_{g}+{r}_{c}
\end{equation}

\begin{equation}
    {r}_{g}=\left\{
        \begin{array}{l}
            {r}_{arrival} \quad \quad \quad \quad \quad \quad \quad \quad \quad if \ {d}_{t} < 0.5 \\
            {\alpha}_{goal} \cdot ({d}_{t}-{d}_{t-1})  \quad \quad \quad \quad \ otherwise
        \end{array}
        \right.
\end{equation}
where ${d}_{t}$ denotes the distance between the UAV and the target point at time t.
\begin{equation}
    {r}_{c}=\left\{
        \begin{array}{l}
            {r}_{collision} \quad \quad \quad \quad \quad \quad \quad \quad \quad \quad if \ crash \\
            {\alpha}_{avoid} \cdot {max}({d}_{safe}-{d}_{min}, 0) \quad otherwise
        \end{array}
        \right.
\end{equation}
We set ${r}_{arrival}=50$, ${r}_{collision}=-10$, ${\alpha}_{goal}=3$, ${\alpha}_{avoid}=-0.05$ and ${d}_{safe}=5$ during the training procedure.

\subsection{Overview}

Following SAC+RAE~\cite{huang2022vision}, we designed a framework based on a variational auto-encoder for visual representation extraction and SAC algorithm for policy learning, as shown in Fig.~\ref{fig:overview}.

In this work, we focus on optimizing the process of representation extraction with causal representation disentanglement to discover hidden causal factors and filter out task-irrelevant (non-causal) factors, which can improve the generalization ability of the DRL model.

Here, following common practices in feature disentanglement~\cite{li2024subspace,kong2022partial,lu2021invariant}, we divide the extracted representations into three typical components as follows:

\begin{itemize}
\item Task-irrelevant components $Z_1 \in \mathbb{R}^{n_1}$
\item Task-relevant but scenarios-specific components $Z_2 \in \mathbb{R}^{n_2}$
\item Task-relevant and scenarios-invariant components $Z_3 \in \mathbb{R}^{n_3}$
\end{itemize}
Obviously, $Z_2$ (\eg, obstacles distribution) and $Z_3$ (\eg, obstacle distance) are causal factors that can provide crucial information for collision avoidance task. Differently, $Z_1$ (\eg, background pattern) is task-irrelevant and can construct spurious correlations due to data-driven training, which results in weak generalization when the UAV system is deployed in unseen scenarios.
Based on our proposed causal representation disentanglement, we can identify these three components and only provide $Z_2$ and $Z_3$ for subsequent policy learning, which explicitly eliminates the influence of $Z_1$.

\subsection{Causal Representation Disentanglement}
The key to causal representation disentanglement is to guide representation components, \ie, $Z_1$, $Z_2$ and $Z_3$, to learn different semantic concepts as expected.
Fig.~\ref{fig:crd} illustrates the procedure of causal representation disentanglement.
In successful practices~\cite{cai2019learning,long2017deep} of feature disentanglement, researchers commonly require data from multiple sources to discover invariant hidden mechanism and identify the causal representation components.
Therefore, in this work, we adopt a simple but effective strategy, \ie, background intervention, to generate multi-domain data.
In addition, we design several critical loss functions to guide representation disentanglement.

\paragraph{Background Intervention}
As shown in our analytical experiment in Fig.~\ref{fig:analysis} and constructed structural causal model (SCM) in Fig.~\ref{fig:scm}, the potentially non-causal factors, \eg, background pattern, can provide some misleading information for collision avoidance that can adversely affect the model's generalization ability to unseen scenarios.
Therefore, to help the network to reveal hidden causal factors, we conduct interventions on non-causal factors to construct multi-domain data by proposing the background intervention strategy.
Here, building on previous works~\cite{lv2022causality,xu2021fourier,oppenheim1981importance}, we utilize an important property of Fourier transform. After Fourier transform, the phase component retains the high-level semantic information of the original signal, \ie, causal factor, while the amplitude component contains the underlying statistical characteristics, \ie, non-causal factors.
Based on this unique property, we conduct random disturbances on the amplitude component of input image $x$:
\begin{equation}
    F(x)=A(x)e^ {-j P(x)}
\end{equation}
\begin{equation}
    \widehat{A}(x)= \lambda A(x)
\end{equation}
\begin{equation}
    F(x^{aug})= \widehat{A}(x)  e^{-j P(x)}
\end{equation}
\begin{equation}
    x^{aug} = F^{-1} (F(x^{aug}))
\end{equation}
where A(x) and P(x) denote the amplitude and phase components respectively, $\lambda$ is a random ratio.
Besides, we also apply several commonly used data augmentation techniques to achieve intervention, \eg, random noise, motion blurring and contrast stretching.

\paragraph{Loss Function}

As shown in Fig.~\ref{fig:crd}, we adopt a variational auto-encoder (VAE)~\cite{kingma2014auto} for representation disentanglement.
% Specifically, the encoder is responsible for generating feature $h$. The bottleneck and $f_\sigma$ , $f_\mu$, utilizes the reparameterization trick to generate the representation $z$. 
VAE is an approximate inference framework based on Bayesian statistics principles, aimed at estimating the posterior distribution in complex models. It models the distribution over the latent space $Z \subseteq \mathbb{R}$ and approximates the true latent space distribution $p(z|x)$ by optimizing the encoder $q(z|x)$. Specifically, the encoder learns the distribution of the latent space $Z$ from the input image $x$ through $q(z|x)$, while the decoder $q(x|z)$ is responsible for reconstructing the sampled point $z \in Z$ from the latent space \ie, the representation, back into an image similar to the input data.
To guide different representation components to learn proper semantic concepts, we design several loss functions.
Firstly, we adopt reconstruction loss to ensure representation $Z$ to encode all visual elements:
\begin{equation}
\mathcal{L}_{vae} = -\mathbb{E}_{q(z|x)}[\log q(x|z)] + D_{KL}[q(z|x) || p(z)]
\end{equation}
where $p(z)$ is the prior distribution.

To optimize the network structure of the bottleneck and increase model diversity, we have restructured $h$:
\begin{equation}
\mathcal{L}_{rec} = (h - h_{rec})^2
\end{equation}

From a causal perspective, the process of background disturbance can be seen as a causal intervention~\cite{pearl2009causality} on images in Fig.~\ref{fig:scm}, which allows us to identify causal factors $S$ within the images.
To separate the scenarios-specific parts within the representation, we design an alignment loss to identify $Z_3$ as follows:
\begin{equation}
\mathcal{L}_ {C}=\frac{1}{C}\sum_{i=1}^{C}||{z}_{3} - {z}_{3,aug}^{i}||^2
\end{equation}
where $C$ represents the total number of types of data augmentation, and ${z}_{3,aug}^{i}$ represents the augmented image obtained through the $i$th type of data augmentation method.

Besides, to ensure the causal components, \ie, $Z_2$ and $Z_3$, is task-relevant and can provide necessary information for policy learning, we pass the sub-features 
$o=[Z_2, Z_3]$ to subsequent policy learning.
Building on previous works~\cite{haarnoja2018soft}, the learning process of the SAC model can generally be summarized in two steps.

The first one is policy evaluation step with the objective to accurately approximate the Q-function:
{\small
\begin{equation}
\mathcal{L}_{Q} = \mathbb{E}_{({o}, {o}^{'}) \sim q(z|x), ({a}, r) \sim \mathcal{B}} \left[ \left( Q({o}, {a}) - r - \gamma \bar{V}({o}^{'}) \right)^2 \right]
\end{equation}
}

In each iteration, $({o}, {o}^{'})$ are sampled from the encoder $q(z|x)$ and $({a},r)$ are sampled from the replay buffer $\mathcal{B}$.
\begin{equation}
    \bar{V}({o}) = \mathbb{E}_{{a} \sim \pi_{\theta}}\left[ \bar{Q}({o},{a}) - r - \alpha \log \pi({a}|{o}) \right]
\end{equation}
where $\bar{Q}$ represents the target Q-function.

The second one is policy improvement step, and the objective is to update the model's policy:
\begin{equation}
    \mathcal{L}_{\pi}=\mathbb{E}_{{o} \sim {q(z|x)}}\left[D_{KL}(\pi(\cdot|{o})\| \mathcal{Q}({o},\cdot))\right] 
\end{equation}
where $\mathcal{Q}({o},\cdot)\propto$ exp $\{\frac{1}{\alpha} Q({o},\cdot)\}$.

Note that only $Z_2$ and $Z_3$ are fed into the reinforcement learning model, and thus the influence of $Z_1$ is eliminated.

\begin{table}[t]
    \vspace{2mm}
    \caption{Hyperparameters for policy trainning.}
    \label{table_1}
    \begin{center}
        \begin{tabular}{p{5cm} p{2cm}}
            \toprule
            Parameters name& Value \\
            \midrule
            \multirow{1}[1]{*}{Replay buffer $\mathcal{B}$ capacity} & 20000 \\
            \multirow{1}[1]{*}{Batch size} & 128 \\
            \multirow{1}[1]{*}{Max episodes} & 300 \\
            \multirow{1}[1]{*}{Update times} & 400 \\
            \multirow{1}[1]{*}{Discount $\gamma$} & 0.99 \\
            \multirow{1}[1]{*}{Optimizer} & Adam \\
            \multirow{1}[1]{*}{Encoder learning rate} & $10^{-4}$ \\
            \multirow{1}[1]{*}{Critic learning rate} & $10^{-4}$ \\
            \multirow{1}[1]{*}{Critic target update frequency} & 2 \\
            \multirow{1}[1]{*}{Critic Q-function soft-update rate $\tau_{Q}$} & 0.01 \\
            \multirow{1}[1]{*}{Critic encoder soft-update rate $\tau_{enc}$} & 0.05 \\
            \multirow{1}[1]{*}{Actor learning rate} & $10^{-4}$ \\
            \multirow{1}[1]{*}{Actor update frequency} & 2 \\
            \multirow{1}[1]{*}{Actor log stddev bounds} & [-10,2] \\                        
            \bottomrule
        \end{tabular}        
    \end{center}
    \label{tab:parameter}
 \end{table}

\section{EXPERIMENT}
\subsection{Experiment and Parameter Setting}

We have developed several simulation environments using Unreal Engine (UE) and Airsim~\cite{airsim2017fsr} simulator. All simulations were carried out on a system running Ubuntu 20.04, equipped with an Intel i7-12700 processor and a NVIDIA GeForce RTX 3060 GPU.

In each simulation round, all UAVs are assigned with randomly selected starting positions and target points within the three-dimensional space. This design ensures that UAVs can thoroughly explore their high-dimensional observational space, thereby enhancing the robustness of the learned strategies. TABLE~\ref{tab:parameter} shows the hyperparameters used in this work.
\begin{figure*}[t]
    \centering
    \includegraphics[width=450pt]{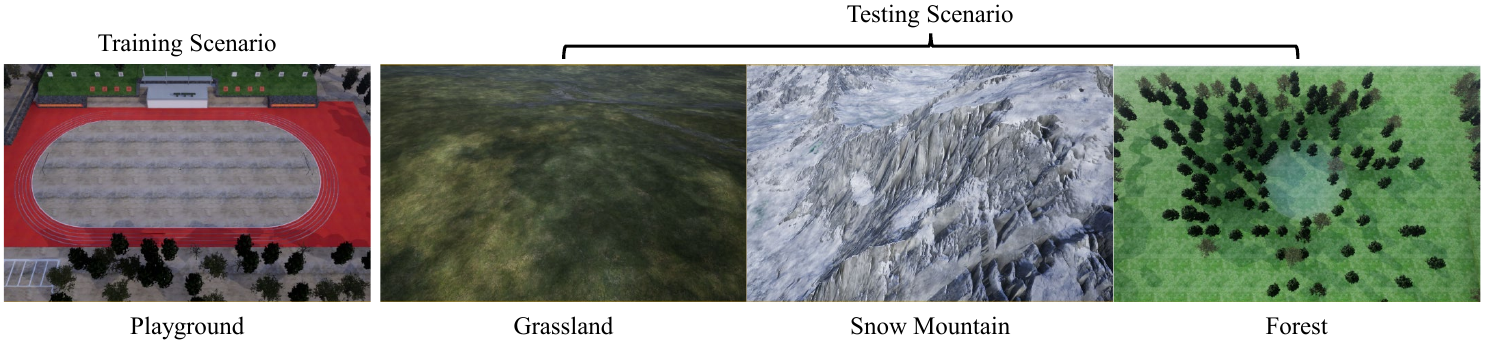}
    %\vspace{2mm}
    \caption{\textbf{Simulation scenarios for model training and testing.} Specifically, playground scenario is used for model training, while grassland, snow mountain and forest scenarios are used for testing.}
    \label{fig:scenarios}
    \vspace{-2mm}
\end{figure*}

\begin{table*}
    \renewcommand{\arraystretch}{1.1}
    \caption{Performance (as mean/std) comparison with the SOTA method in unseen testing scenarios under random initialization.}
    \label{tab:performance_random}
    \begin{center}
        \begin{tabular}{cccccc}
            \toprule
            Scenario& Method & Success Rate (\%) & SPL (\%) & Extra Distance (m) & Average Speed (m/s)\\
            \midrule
            \multirow{2}[1]{*}{Grassland} & SAC+RAE & 85.8 & 74.2 & 1.515/1.286 & 0.899/0.136 \\ 
                  & Our method & \textbf{88.6} ($\uparrow 2.8$)  & \textbf{82.3} ($\uparrow 8.1$)& \textbf{0.725/0.951} & \textbf{0.991/0.105}\\                  
            \hline
            \multirow{2}[1]{*}{Snow Mountain} & SAC+RAE & 80.6 & 66.7 & \textbf{2.091/1.685} & 0.815/0.178 \\
                  & Our method & \textbf{88.4} ($\uparrow 7.8$)  & \textbf{67.9} ($\uparrow 1.2$)& 3.066/2.657 & \textbf{0.962/0.076} \\                  
            \hline
            \multirow{2}[1]{*}{Forest} & SAC+RAE & 73.1 & 60.3 & \textbf{2.125/1.715} & 0.806/0.174 \\
                  & Our method & \textbf{88.9} ($\uparrow 15.8$)  & \textbf{72.2} ($\uparrow 11.9$)& 2.600/2.788 & \textbf{0.984/0.121} \\                  
            \bottomrule
            \end{tabular}
    \end{center}
    \vspace{-2mm}
 \end{table*}

\begin{table*}
    \renewcommand{\arraystretch}{1.1}
    \caption{Performance (as mean/std) comparison with the SOTA method in unseen testing scenarios under circle initialization.}
    \label{tab:performance_circle}
    \begin{center}
        \begin{tabular}{cccccc}
            \toprule
            Scenario& Method & Success Rate (\%) & SPL (\%) & Extra Distance (m) & Average Speed (m/s)\\
            \midrule
            \multirow{2}[1]{*}{Grassland} & SAC+RAE & 53.8& 50.1& 1.769/0.511& 0.804/0.049\\
                  & Our method & \textbf{98.7} ($\uparrow 44.9$)& \textbf{92.7} ($\uparrow 42.6$)& \textbf{1.549/0.335}& \textbf{1.122/0.058}\\                 
            \hline
            \multirow{2}[1]{*}{Snow Mountain} & SAC+RAE & 72.4 & 67.8 & \textbf{1.655/0.882} & 0.959/0.141 \\ 
                  & Our method & \textbf{99.5} ($\uparrow 27.1$)& \textbf{92.1} ($\uparrow 24.3$)& 1.692/0.887& \textbf{1.046/0.065}\\                  
            \hline
            \multirow{2}[1]{*}{Forest} & SAC+RAE & 40.9 & 32.0 & \textbf{6.677/1.296} & 0.970/0.141 \\ 
                  & Our method & \textbf{99.4} ($\uparrow 58.5$)& \textbf{71.0} ($\uparrow 39.0$)& 9.955/3.827 & \textbf{1.009/0.054} \\                  
            \bottomrule
            \end{tabular}
    \end{center}
    \vspace{-2mm}
 \end{table*}

\subsection{Performance Metrics and Experiment Scenarios}

\paragraph{Performance Metrics}
Following SAC+RAE~\cite{huang2022vision}, to assess the performance across different scenarios, we employed the following evaluation metrics:
\begin{itemize}
\item Success Rate:
The percentage of agents that reach their own target points within the specified time without any collisions.
\item SPL (Success weighted by Path Length):
Measuring whether a UAV can successfully reach its destination via the shortest path:
\begin{equation}
    SPL = \frac{\sigma\cdot l}{max(l,p)} 
\end{equation}

where $l$ represents the shortest-path distance from the UAV's starting point to the target point, while $p$ denotes the actual path length taken by the UAV. $\sigma$ is a binary indicator of success or failure.
\item Extra Distance:
The average extra distance of UAVs' journey compared to the shortest-path distance between the starting point and the target point.
\item Average Speed:
The average speed of all UAVs throughout testing.
\end{itemize}

\paragraph{Scenarios}

We have developed four distinct scenarios for evaluating the generalization ability of DRL models, where one for training and the others for testing, as shown in Fig.~\ref{fig:scenarios}. 
Specifically, we train the model in a common playground scenario. And considering UAV practical applications like geological survey~\cite{tziavou2018unmanned}, search and rescue~\cite{ashour2023applications}, we design three typical scenarios for testing, \ie, grassland, snow mountain and forest.

\paragraph{Initialization}
Following SAC+RAE, we consider two typical initialization patterns of the starting and target points for each UAV.
\begin{itemize}
\item Random Pattern: The starting and target positions of each UAV are randomly generated within a cubic area measuring 16 meters in length and width, and 4 meters in height.
\item Cycle Pattern: All UAVs are uniformly initially positioned on a circle with a radius of 12 meters at a specific height. Their target points are set on the opposite side of the circle.
\end{itemize}

Our approach was exclusively trained in the playground scenario using random pattern initialization, and then evaluated in other unseen scenarios.

\begin{figure*}[t]
    \centering
    \includegraphics[width=350pt]{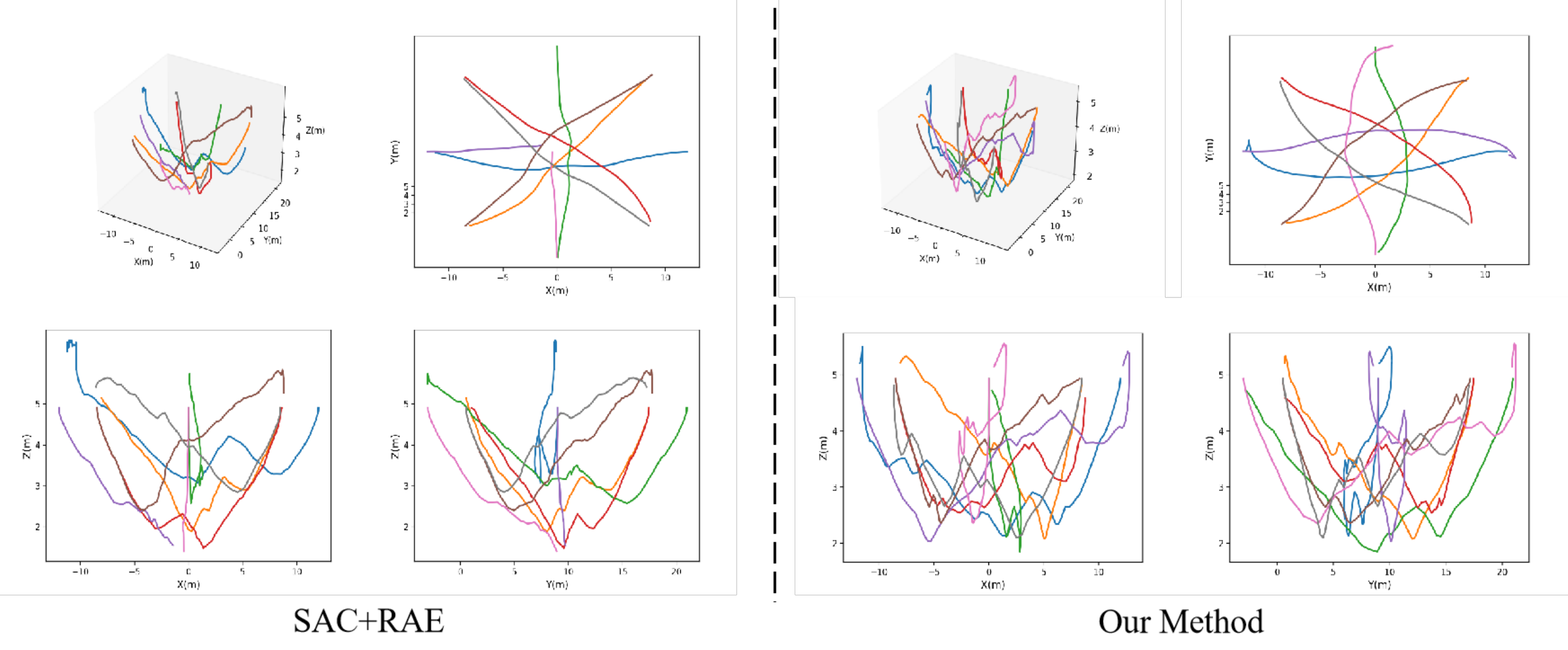}
    %\vspace{2mm}
    \caption{\textbf{Visualization of UAV trajectories in perspective drawing and three-view drawing.} The trajectories of different UAVs are represented by different colors. Best viewed in color.}
    \label{fig:visualization}
    \vspace{-2mm}
\end{figure*}

 \begin{table*}[t]
    \renewcommand{\arraystretch}{1.1}
    \caption{Performance (as mean/std) comparison with the SOTA method with different numbers of UAVs.}
    \label{tab:scale}
    \begin{center}
        \begin{tabular}{cccccc}
            \toprule
            Number& Method & Success Rate (\%) & SPL (\%) & Extra Distance (m) & Average Speed (m/s)\\
            \midrule
            \multirow{2}[1]{*}{6} & SAC+RAE & 77.7 & 64.5 & \textbf{2.059/1.666} & 0.842/0.178 \\
                  & Our method & \textbf{91.0} ($\uparrow 13.3$)  & \textbf{73.9} ($\uparrow 9.4$) & 2.602/2.829 & \textbf{0.997/0.132}\\                  
            \hline
            \multirow{2}[1]{*}{8} & SAC+RAE & 73.1 & 60.3 & \textbf{2.125/1.715} & 0.806/0.174 \\
                  & Our method & \textbf{88.9} ($\uparrow 15.8$)  & \textbf{72.2} ($\uparrow 11.9$) & 2.600/2.788 & \textbf{0.984/0.121} \\
                                   
            \hline
            \multirow{2}[1]{*}{10} & SAC+RAE & 72.8 & 58.8 & \textbf{2.386/1.825} & 0.771/0.175 \\
                  & Our method & \textbf{84.4} ($\uparrow 11.6$)  & \textbf{66.8} ($\uparrow 8$) & 3.006/3.370 & \textbf{0.940/0.112} \\
                  
            \hline
            \multirow{2}[1]{*}{12} & SAC+RAE & 71.1 & 56.8 & \textbf{2.547/1.968} & 0.717/0.163 \\ 
                  & Our method & \textbf{81.9} ($\uparrow 10.8$)  & \textbf{64.7} ($\uparrow 7.9$) & 2.989/2.994 & \textbf{0.905/0.095} \\
            \bottomrule
            \end{tabular}                                                    
    \end{center}
    \vspace{-2mm}
 \end{table*}

\begin{table}[t]
  \begin{center}
    \caption{Ablation study on combination of representation components.}
    \label{tab:component}
    \begin{tabular}{c|c|c|c}  
      \toprule
            $Z_1$ & $Z_2$ & $Z_3$ & Success Rate (\%) \\
            \midrule            
              & \checkmark  & \checkmark & \textbf{88.9} \\
              &  \checkmark & & 64.7\\
              &  & \checkmark & 60.7\\              
              \checkmark & \checkmark & \checkmark & 71.4 \\
            \bottomrule
    \end{tabular}
  \end{center}
  \vspace{-5mm}
\end{table}

\subsection{Performance Comparison}

To evaluate the generalization ability, we tested our method and the previous SOTA method, \ie, SAC+RAE~\cite{huang2022vision}, in three unseen testing scenarios under two initialization settings, as shown in TABLE~\ref{tab:performance_random} and TABLE~\ref{tab:performance_circle}.
From the results, we have the following two observations.
First, our method can achieve significant improvement on navigation success rate and SPL with higher average speed, which clearly shows that the proposed causal representation disentanglement can effectively improve the generalization ability of DRL model in unseen scenarios.
Second, the flight path planned by our method is slightly longer than that of SAC+RAE, because our planed path contains more collision avoidance actions while path planed by SAC+RAE causes more collisions due to the interference from unseen backgrounds.

\subsection{Ablation Study}
In this subsection, we conduct experiments to reveal the effectiveness of our proposed method. All experiments are particularly conducted in the forest scenario under random initialization by default.

\subsubsection{Representation Components}

The key behind our method on improving generalization ability is disentangling representation into three components (\ie, $Z_1$, $Z_2$ and $Z_3$) and only using the task-relevant components (\ie, $Z_2$ and $Z_3$) for subsequent policy learning, which explicitly eliminate the influence of task-irrelevant component (\ie, $Z_1$).
To better reveal the effect of representation components, we conduct ablation study on different combination of components.
As shown in TABLE~\ref{tab:component}, we have two observations as follows.
First, $Z_2$ and $Z_3$ are both important for optimal policy learning and a combination of them can achieve the best performance.
Second, if we use $Z_1$, the generalization ability would be reduced since $Z_1$ contains task-irrelevant information that can easily construct spurious correlations and result in wrong action prediction.

\subsubsection{Scalability}

To assess the scalability of our method, we adjust the number of UAVs during testing (\ie, 6, 8, 10 and 12), where the model is still trained with 8 UAVs.
As shown in TABLE~\ref{tab:scale}, our method can achieve consistent performance improvement over SAC+RAE, which demonstrates that our method can better generalize to multi-UAV system with different scale.

\subsubsection{Trajectory Visualization}

Additionally, we visualize the trajectories of UAVs under circular initialization, as shown in Fig.~\ref{fig:visualization}. Here, trajectories are shown in both perspective drawing and three-view drawing, where trajectories of different UAVs are drawn by different colors.
Our method shows smoother and more complete trajectories while SAC+RAE results in collisions of 2 UAVs, which demonstrates the improved ability of our method for robust and effective path planning.

\section{CONCLUSIONS}

%In this work, we propose a novel representation learning method, \ie, causal representation disentanglement, to improve the generalization ability of DRL techniques in unseen scenarios.
In this work, we study the generalization issue of current DRL-based multi-UAV collision avoidance system in unseen scenarios.
To address this issue, we design a novel representation learning method, \ie, causal representation disentanglement, to divide the visual representation into several components with specific semantic concepts and only pass those causal components for subsequent policy learning, which can effectively eliminate the influence of the task-irrelevant component and thus improve the model's generalization ability.
Extensive experiments on multiple unseen testing scenarios validate the feasibility and effectiveness of the proposed method, which outperforms the previous state-of-the-art method.


%%%%%%%%%%%%%%%%%%%%%%%%%%%%%%%%%%%%%%%%%%%%%%%%%%%%%%%%%%%%%%%%%%%%%%%%%%%%%%%%

%%%%%%%%%%%%%%%%%%%%%%%%%%%%%%%%%%%%%%%%%%%%%%%%%%%%%%%%%%%%%%%%%%%%%%%%%%%%%%%%

%%%%%%%%%%%%%%%%%%%%%%%%%%%%%%%%%%%%%%%%%%%%%%%%%%%%%%%%%%%%%%%%%%%%%%%%%%%%%%%%
%\section*{APPENDIX}


%\section*{ACKNOWLEDGMENT}

%The preferred spelling of the word ÒacknowledgmentÓ in America is without an ÒeÓ after the ÒgÓ. Avoid the stilted expression, ÒOne of us (R. B. G.) thanks . . .Ó  Instead, try ÒR. B. G. thanksÓ. Put sponsor acknowledgments in the unnumbered footnote on the first page.

%%%%%%%%%%%%%%%%%%%%%%%%%%%%%%%%%%%%%%%%%%%%%%%%%%%%%%%%%%%%%%%%%%%%%%%%%%%%%%%%

% \newpage
\bibliographystyle{IEEEtran}
\small\bibliography{reference}

\end{document}